\pdfoutput=1  
\PassOptionsToPackage{table}{xcolor} 
\documentclass{article}

\usepackage[preprint]{corl_2026} 

\usepackage{booktabs}   
\usepackage{multirow}   
\usepackage{geometry}   
\usepackage{graphicx}
\usepackage{amsmath}
\usepackage{dsfont}
\usepackage{wrapfig}
\usepackage{listings}   
\newcommand{\Cost}{C}
\newcommand{\Prompt}{l}
\newcommand{\Image}{I}
\newcommand{\Pointcloud}{P}
\newcommand{\Policy}{\pi}
\newcommand{\Dict}{D}

\title{G2-Nav: Grounded and Guarded Vision-Language Costmaps for Robot Social Navigation}

\author{
  Yuwen Liao\textsuperscript{1}, Yihang Lan\textsuperscript{1}, Yizhuo Yang\textsuperscript{1}, Ruimeng Liu\textsuperscript{1},\\
  \bfseries Xinhang Xu\textsuperscript{1}, Shenghai Yuan\textsuperscript{1}, Lihua Xie\textsuperscript{2}\\
  \textsuperscript{1}\,Nanyang Technological University, Singapore 639798\\
  \parbox[t]{0.9\textwidth}{\centering
    \textsuperscript{2}\,NTU--VinUni Joint Research Laboratory for Robotics and Embodied AI, \\
    School of Electrical and Electronic Engineering, Nanyang Technological University, Singapore 639798 and VinUniversity, Hanoi 100000, Vietnam}
}

\begin{document}
\maketitle


\begin{abstract}
Social navigation requires the robot to reason and respond in complex real-world environments. 
While recent works attempt to incorporate human-level intelligence into robot planning using large Vision-Language Models (VLMs), end-to-end frameworks often create an unpredictable black-box, and existing instruction-following methods are not designed for full autonomy.
To bridge this gap, we present G2-Nav, a novel framework that grounds abstract social reasoning and guards safe real-world deployment.
Instead of asking the VLM for direct planning decisions, G2-Nav translates its semantic reasoning into a vision-language costmap with reliability and interpretability.
The VLM evaluates traversable regions and social agents from open-set perception, mapping social context into the costmap.
To improve real-world robustness, the VLM performs semantic verification on upstream tracking, and we introduce a high-frequency safety check to guard against system latency prior to trajectory generation.
We demonstrate through real-world experiments that G2-Nav delivers safe, efficient, and socially compliant autonomous navigation in unstructured environments. 
Code is available at https://github.com/centiLinda/G2-Nav.

\end{abstract}




\section{Introduction}

\begin{wrapfigure}{r}{0.37\textwidth}
    \vspace{-12pt}
    \centering
    \includegraphics[width=\linewidth]{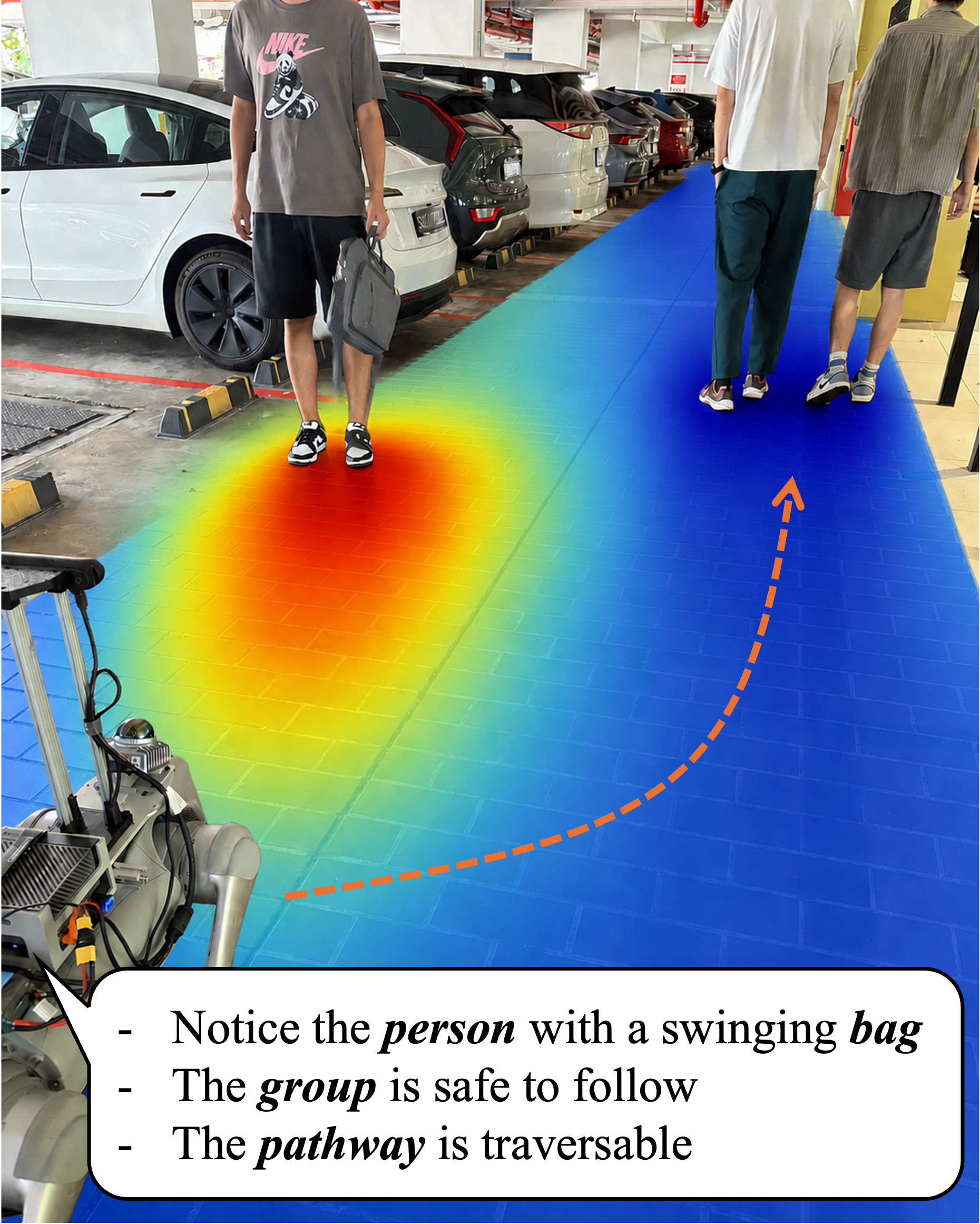}
    \vspace{-16pt}
    \caption{Grounding social context.}
    \label{fig:illustration}
    \vspace{-10pt}
\end{wrapfigure}

Understanding social context is crucial for autonomous robots navigating in human-centric environments.
Beyond basic obstacle avoidance, early works rely on hand-crafted behavioural rules~\cite{helbing1995social} and assumptions~\cite{van2011reciprocal} that struggle to capture dynamic human interactions and unspoken social norms.
While subsequent works have proposed various data-driven techniques to learn from human demonstrations~\cite{korbmacher2022review, nguyen2023toward}, many of the latest works have pivoted towards large Vision Language Models (VLMs), leveraging their internet-scale semantic reasoning~\cite{kawaharazuka2025vision}.
However, directly deploying VLMs for trajectory and action generation results in a black-box system that lacks safety assurance~\cite{huang2025trust}.
Alternatively, using VLMs as evaluators for sampled trajectories may help to reduce numerical hallucinations, but sampling efficiency and diversity become a severe bottleneck~\cite{li2026vision}.
We therefore ask the question: 
\textit{How to effectively and safely ground VLM social reasoning into reliable real-world robot navigation?}

In this paper, we look into the costmap, a foundational representation in robot navigation that can serve as a mathematically sound and interpretable interface for abstract semantic reasoning.
Traditional trajectory optimisation methods mainly treat occupancy grid as the costmap for obstacle avoidance~\cite{163777}, or assign pre-defined costs to humans~\cite{ge2002dynamic, papadakis2013social}, which ignore more diverse social interactions.
Recent works in vision-language navigation have utilised costmaps for instruction following, where detailed human commands, such as "stay on the concrete and follow the person wearing black", are translated into a heatmap and passed to downstream controllers~\cite{weerakoon2025behav, yuan2025opennav, huang2023visual}.
However, these high-level and detailed instructions are designed for specific scenarios, and not suitable for autonomous robots.
We argue that a unique capability of VLMs in social navigation is to identify and analyse interested agents from complex and unstructured real-world environments, as illustrated in Fig.~\ref{fig:illustration}.
We should explore the best of both worlds by analysing vibrant social cues using VLMs and grounding it using costmaps for reliable robot behaviours.

To this end, we propose \textbf{G2-Nav}, a novel social navigation framework that \textbf{G}rounds social context reasoning and \textbf{G}uards safe real-world deployment.
We first perform open-set perception to obtain a basic understanding of surrounding objects and their spatial locations.
Instead of treating all as equally dangerous hazards, we use a pretrained VLM to differentiate between ground regions, interested social agents, and irrelevant objects.
The VLM will then identify the traversable region and describe the intended interaction with each social agent using a scoring system, which can be effectively mapped to a bounded cost space.
We observe that birds-eye-view (BEV) object tracking can be disrupted by noisy sensor data in the wild, therefore we further utilise the VLM's semantic reasoning to verify upstream perception plausibility.
To account for VLM response latency, we apply a safety layer to check for immediate danger within the robot's reflex zone, before generating finalised robot actions using local controllers.

Our contributions are threefold.
First, we propose a novel social context representation by grounding VLM reasoning into a vision-language costmap.
Second, we design robust mechanisms including upstream verification to account for noisy tracking, and a safety reflex zone to guard against system latency.
Third, we demonstrate through extensive experiments that the proposed G2-Nav framework promotes social compliance while preserving safety and efficiency in real-world navigation.


\section{Related Work}

Early works in robot navigation consider social context as pre-defined rules and assumptions, such as equal collision avoidance responsibility~\cite{van2011reciprocal}, repulsive forces~\cite{helbing1995social}, and various types of social zones~\cite{ge2002dynamic, papadakis2013social}.
Human motions and interactions have also been modelled using more advanced neural networks~\cite{liu2025height}, but they are still limited to modelling humans as dynamic obstacles, ignoring other types of interactions and scene semantics.
Social context can also be implicitly embedded into end-to-end frameworks using reinforcement learning and imitation learning~\cite{shah2023vint, sridhar2024nomad, bar2025navigation, liu2025citywalker}, but they tend to experience domain gaps when deployed in different environments.

To achieve zero-shot social understanding, more recent works use VLMs as the core cognitive module.
Similar to previous end-to-end frameworks, VLMs can be finetuned to directly generate robot trajectories and actions~\cite{zhou2024navgpt, chen2026socialnav, song2024vlm}.
However, VLMs are often observed to produce numerical hallucinations, giving unstable responses between steps or randomly injecting dangerous actions~\cite{sathyamoorthy2024convoi, huang2025trust}.
Other works argue that VLMs are better suited for answering \textit{yes-or-no} questions rather than open-ended \textit{how} questions~\cite{nasiriany2024pivot}. 
Therefore, these methods use VLMs to select the most socially compliant trajectory or action from a candidate set produced by external samplers~\cite{fang2026obstacles, song2025vl, potnis2026catnav}. 
However, this brings a difficult trade-off between sample quantity and state-space coverage~\cite{li2026vision}.

Of relevance to our work is the literature on incorporating VLMs to generate navigation costmaps.
Most of these works rely on detailed instructions, covering only selected agents and navigation behaviours~\cite{weerakoon2025behav, yuan2025opennav, huang2023visual, chen2025lisn}.
These instructions are designed for specific scenarios and are not available in an autonomous setting.
Humans are also treated as equally dangerous safety hazards, unless otherwise specified in the tasks~\cite{chen2025lisn, zhang2025socialnav}.
CORE~\cite{ravichandran2026contextual} aligns with our idea of identifying interested objects and traversable regions from unstructured environments, but the application scenario is for exploring indoor structures, which lacks dynamic social interactions.
Vi-LAD~\cite{elnoor2025vi} also uses VLM to perform semantic reasoning, but the task is simply selecting a general direction where the crowd may appear.
In comparison, G2-Nav is more suitable for real-world robot deployment in human-centric social environments.

\section{Methodology}

\begin{figure} [t]
    \centering
    \includegraphics[width=\linewidth]{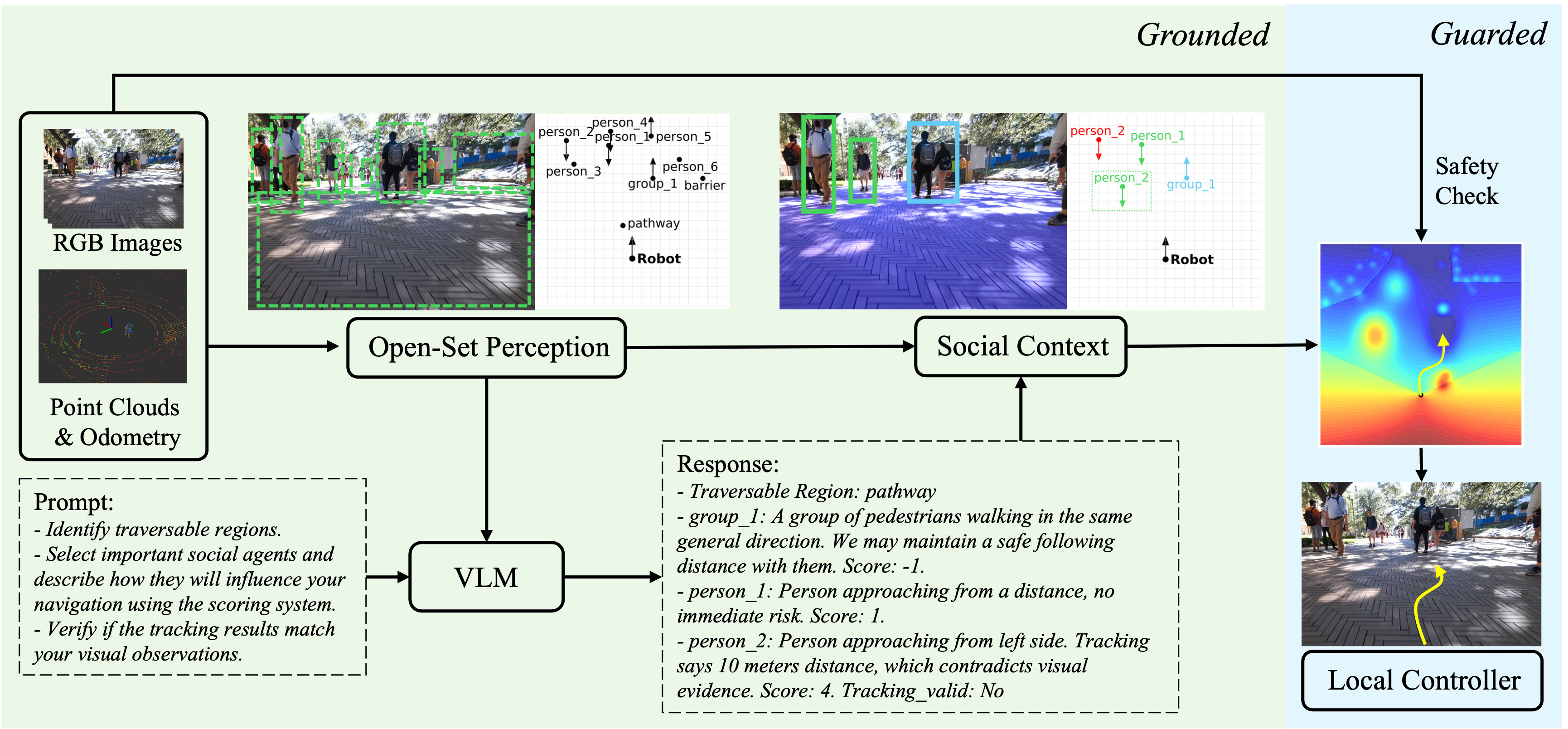}
    \caption{Overview of the G2-Nav framework. In this example, the VLM suggests that person\_2 is tracked at the wrong location, which is then corrected by a depth re-registration. The safety check captures a person stepping in from the right, which is reflected in the final costmap.}
    \label{fig:framework}
\end{figure}

The proposed G2-Nav framework begins by processing sensor data with the open-set perception module, which detects visible objects along with their locations and motions (Sec.~\ref{sec:perception}).
The VLM analyses this perception information, extracts social cues by identifying traversable regions and scoring relevant objects, and additionally verifies the perception output against visual input (Sec.~\ref{sec:grounding}).
These social contexts can then be formulated into a costmap with additional safety checks (Sec.~\ref{sec:guarding}), and be used to generate robot control actions.
The overall framework is illustrated in Fig.~\ref{fig:framework}.

\subsection{Problem Formulation}
The social navigation task can be formulated as a mapping between raw sensor data and robot actions.
At time $t$, the robot receives a sequence of RGB images $\Image^{t-n:t} = \{\Image^{t-n}, \dots, \Image^t\}, \text{ where } \Image \in \mathds{R}^{H \times W \times 3}$ and $n$ is the number of historical observations, LiDAR point clouds $\Pointcloud^t$, and a target location $g^t \in \mathds{R}^{2}$ in the robot odometry.
A navigation policy $\Policy(a^t \mid \Image^{t-n:t}, \Pointcloud^t, g^t)$ generates action $a^t$ from the action space $\mathcal{A}$.
We follow existing works~\cite{shah2023vint, shah2023gnm} to use trajectory waypoints as the action space, so that our framework is embodiment-agnostic.
During deployment, a robot-specific controller is used to track the trajectories considering dynamics constraints.
Time superscripts will be omitted when no confusion is aroused.

\subsection{Open-Set Perception}
\label{sec:perception}
Given raw sensor data $\Image$ and $\Pointcloud$, we obtain surrounding objects' spatial information through open-set recognition, detection, and tracking.
To maintain rich context details, we first perform open-set object recognition using RAM++~\cite{zhang2024recognize} to get a list of text tags of all visual elements in the current image $\Image$.
This includes both ground regions (e.g., grass, pathway) and objects (e.g., person, backpack).
We then perform object detection using GroundingDINO~\cite{liu2024grounding} to associate the text tags with 2D bounding boxes and assign each object with a unique ID $o_{i}$.
Depth registration is done by projecting the 2D bounding box into the point cloud, and the object is identified using DBSCAN~\cite{ester1996density} clustering.
We then use Kalman Filter to maintain a list of tracked objects across timesteps.
Formally, we define a state dictionary to hold each object's information $\Dict = \{ o_i \mapsto (b_{i}, x_{i}, v_{i}) \}$, where $b_{i}$ is 2D bounding box, $x_{i}$ is location in robot's odometry, and $v_{i}$ is velocity estimated from consecutive frames.


A major gap in the depth registration step is associating with the wrong point cloud cluster.
We will address this in Sec.~\ref{sec:grounding} through VLM upstream verification.
Although there are other ways to obtain object locations, such as directly from point clouds~\cite{zhou2018voxelnet} or image-based depth estimation~\cite{yang2024depth}, these methods generalise poorly on real-world data collected from robot onboard sensors.

\subsection{Grounding Social Context}
\label{sec:grounding}
Given preliminary object information $\Dict$, image histories $\Image^{t-n:t}$, and a prompt $\Prompt$, we use a pretrained VLM to extract social cues in a formatted manner, which can then be grounded into the vision-language costmap.
Specifically, we design three prompt elements to elicit VLM semantic reasoning in three unique directions.

\noindent\textbf{Traversability Identification.} 
Among the open-set detections, VLM needs to identify elements that belong to different types of ground regions.
The VLM will further identify a traversable region, and save other non-traversable regions in a separate list.
These ground regions are only used for traversability analysis and will not be treated as physical obstacles.
The traversable region and non-traversable region list are queried and updated at every step to account for changing environments.
We use SAM~\cite{kirillov2023segment} segmentation model to obtain a traversable mask $M \in \{0, 1\}^{H \times W}$.

\noindent\textbf{Social Scoring.} 
Apart from ground regions, the VLM is also asked to remove irrelevant objects from $\Dict$, such as those that are far away, or unlikely to have any interaction with the robot.
The rest of the objects $\{o_j\} \subseteq \{o_i\}$ are assigned a social score.
Objects that pose immediate danger are given higher scores, while objects that simply need to be monitored are given lower scores.
Inspired by a recent work~\cite{liao2025following}, humans who travel in similar directions as the robot can be treated as social leaders.
They provide strong cues on unspoken norms, such as keeping to which side of the street, or avoiding certain areas that are not visibly marked.
Therefore, we also ask the VLM to assign a special negative score to humans that could provide this implicit guidance to the robot.
To account for VLM response latency, recently detected objects are assigned a default score until they are analysed by the VLM.
This \textit{asynchronous} design ensures that the robot performs real-time planning without being bottlenecked by the VLM inference cycle.

\noindent\textbf{Upstream Verification.} 
As mentioned in Sec.~\ref{sec:perception}, the noisy point cloud collected by onboard LiDAR can cause confusion during depth registration.
Complex background and floating particles are sometimes mistaken as the object when searching in the projected bounding box, and unreliable locations will then lead to wrong velocity estimation.
Therefore, we ask the VLM to verify if the object's depth and heading match with visual observations $\Image^{t-n:t}$ (e.g., "Does this car look like 20 meters away?", "Is this person moving towards us?").
If there is a depth mismatch, we will perform depth registration again while filtering out the previous false cluster.
If there is a heading mismatch, we will increase the object's social score to account for its uncertain motion.
Detailed examples can be found in the supplementary material.

The overall social context is represented in a combination of the traversable mask $M$ and social agent information $\Dict^* = \{ o_j \mapsto (b_{j}, x_{j}, v_{j}, s_{j}) \}$, where $s_j$ is the social score.

\subsection{Safeguarded Costmap Generation}
\label{sec:guarding}
We formulate the BEV costmap as $\Cost : \mathcal{W} \rightarrow \mathds{R}$, where $\mathcal{W} = [-X, X] \times [-Y, Y]$ is the planning horizon in robot odometry, defined as a weighted combination of social context and standard navigation terms:

\begin{equation}
    \Cost = \lambda_1 \Cost_{\text{goal}} + \lambda_2 \Cost_{\text{obs}} + \lambda_3 \Cost_{\text{obj}} + \Cost_{\text{trav}},
\end{equation}

where $\lambda_{1:3}$ are weights for each individual cost component, defined as follows:
\begin{itemize}
    \item Goal Attraction ($\Cost_{\text{goal}}$) penalises the Euclidean distance from current location to goal $g$.
    \item Static Obstacles ($\Cost_{\text{obs}}$) is the occupancy map projected from point cloud $\Pointcloud$, inflated by robot radius.
    \item Social Agent ($\Cost_{\text{obj}}$) models social interaction as a sum of elliptical Gaussians:
    \begin{equation}
        C_{\text{obj}} = \sum_j s_j \cdot \mathcal{N}(x_j, \Sigma(v_j)). 
    \end{equation}
    \item Traversability Mask ($\Cost_{\text{trav}}$) penalises non-traversable regions outside the binary mask $M$.
\end{itemize}

To safeguard against fast-moving obstacles that are not registered by the perception module nor by the VLM, we implement a high-frequency reflex check.
We define a dynamic reflex zone along the robot's predicted trajectory (detailed definitions can be found in the supplementary material).
LiDAR points falling within this zone are checked against the current dictionary of objects. 
Points belonging to a known object (e.g., a static, harmless fence) are ignored, while unassigned points trigger a reflex response, injecting high penalties into the costmap, forcing immediate robot response.

\subsection{Local Controller}
\label{sec:control}
Given the costmap $\Cost$, we use Gradient Descent to generate the trajectory and Pure Pursuit to track trajectory waypoints considering robot dynamics constraints.
Although more sophisticated pathfinding and trajectory tracking methods can be used, the rich semantic and spatial information embedded in the vision-language costmap makes these classic tools highly effective in our experiments.

\section{Experiments}

We evaluate the performance of G2-Nav in two different settings: 
1) \textbf{Recorded} social navigation dataset SCAND~\cite{karnan2022socially} with human expert demonstration, 
2) \textbf{Interactive} deployment in a crowded campus environment.
The recorded open-loop comparison and interactive closed-loop validation complement each other, 
and both showcase real-world human-centric social interactions.
\begin{wrapfigure}{r}{0.4\textwidth}
    \vspace{+23pt}
    \centering
    \includegraphics[width=\linewidth]{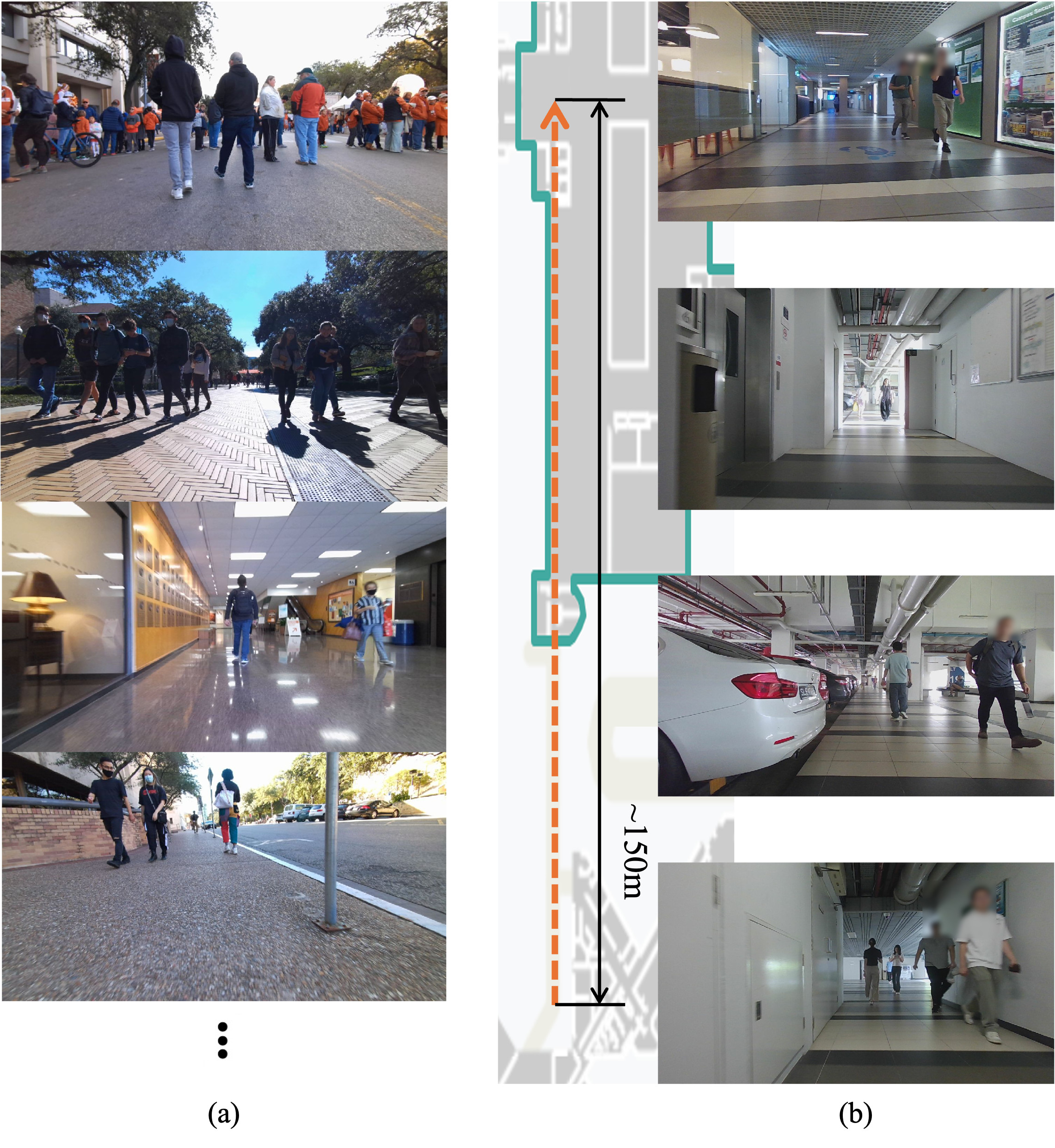}
    \vspace{-18pt}
    \caption{Example scenarios from recorded (a) and interactive (b) settings.}
    \label{fig:scenes}
    \vspace{-20pt}
\end{wrapfigure}

\vspace{-10pt}
\subsection{Experiment Settings}


\noindent\textbf{Scenarios.}
SCAND originally consists of 138 trials of robot onboard data.
We remove trials that are too short, missing topics, and those that involve stair-climbing, which yields 83 trials and a total duration of 5.6 hours.
We conduct the real-world experiment in a 150m-long crowded campus corridor connecting academic buildings and carparks.
During each data collection, we make sure to run all the methods so they are tested with similar crowd densities.
We collect 6 trials for each method.

\noindent\textbf{Implementation Details.}
For the recorded setting, we use QWen3.5-2B~\cite{qwen3.5} as the local VLM model, served with vLLM on two RTX2080 GPUs with an average query time of 1.5s.
For the interactive setting, due to limited computational resources on the edge device, we use GPT5.4-Nano as the online VLM model with an average query time of 4s.
We use Unitree Go2-W as the robot platform and set the speed limit to $1.0 \text{m/s}$.
The ROS system runs on Jetson AGX Orin, connected to an RGB camera and Livox Mid-360 LiDAR.
Robot odometry is obtained by Super-LIO~\cite{wang2026super}.
We find the set of costmap weights $\lambda_{1:3}$ through grid search using the longest trial in SCAND, and use the same values in both experiment settings. 
Social score $s_j$ is set to $-1$ for leaders, 1 to 5 for the rest, 3 being the default.
The rest of the parameters can be found in the code repository.

\noindent\textbf{Evaluation Metrics.}
For the recorded setting, we compare the similarity between planned trajectory with human-operated ground-truth using L2 Distance (\textbf{L2}) and Maximum Average Orientation Error (\textbf{MAOE}), proposed by CityWalker~\cite{liu2025citywalker}.
For the interactive setting, we use commonly adopted metrics~\cite{francis2025principles} including Safety Violation Time (\textbf{SVT}) for the time spent within 0.3m of any obstacle, Average Time (\textbf{T}\textsubscript{\textbf{avg}}) and Average Distance (\textbf{D}\textsubscript{\textbf{avg}}) taken to reach the goal.
To avoid hardware damage and ensure safety, the human supervisor applies 1) Safety pause when the robot is on collision course (clearance below $\sim$0.1m), 2) Boundary pause when the robot attempts to enter the off-limits carpark.
We report the total \textbf{Intervention} counts for both situations.
The pause period is not counted in SVT and T\textsubscript{avg}.

\noindent\textbf{Baselines.}
We compare G2-Nav with the following navigation methods with different levels of social context understanding: 
1) Dynamic Window Approach (\textbf{DWA})~\cite{fox1997dynamic} performs point cloud-based obstacle avoidance without considering semantic or object-level information, 
2) Social Force (\textbf{SF})~\cite{helbing1995social}\footnote{We use the open-source PySocialForce implementation with default parameters: \url{https://github.com/yuxiang-gao/PySocialForce}.} considers heuristic forces such as goal attraction and neighbours repulsion, 
3) PeopleAsPlanner (\textbf{PAP})~\cite{liao2025following} follows human leaders to navigate through crowds, 
4) \textbf{CityWalker}~\cite{liu2025citywalker} is an end-to-end navigation framework trained using web-scale human demonstration videos, 
5) \textbf{path-etiquette}~\cite{fang2026obstacles} uses VLM to select from sampled trajectories. We use the same VLM model as ours when deploying path-etiquette.
Since DWA, SF, and PAP are reactive planners that generate single-step robot action, they are not tested in the recorded setting since the evaluation requires planned trajectories.
For CityWalker and path-etiquette, we use the same Pure Pursuit algorithm in the interactive setting for trajectory tracking.
\begin{wraptable}{r}{0.55\textwidth}
    \centering
    \vspace{+28pt}
    \begin{tabular}{l cc}
        \toprule
        \textbf{Method} & \textbf{L2 (m) $\downarrow$} & \textbf{MAOE ($^\circ$) $\downarrow$} \\ 
        \midrule
        CityWalker~\cite{liu2025citywalker}      & 0.60 & 16.81 \\
        path-etiquette~\cite{fang2026obstacles}  & 0.97 & 20.14 \\
        \midrule
        \textbf{G2-Nav} (ours)                   & \underline{0.55} & \textbf{15.64} \\
        \quad w/o traversability                 & 0.84 & 18.10 \\
        \quad w/o scoring                        & 1.12 & 21.08 \\
        \quad w/o verification                   & 0.57 & 16.37 \\
        \quad w/o safeguard                      & \textbf{0.48} & \underline{15.90} \\
        \bottomrule
    \end{tabular}
    \caption{Quantitative comparisons under recorded setting. Results are averaged from 83 trials. The best performance is in \textbf{bold} and the second-best is \underline{underlined}.}
    \label{tab:quantitative_scand}
    \vspace{-18pt}
\end{wraptable}

\vspace{-13pt}
\subsection{Recorded Setting}

\noindent\textbf{Quantitative Analysis.}
As shown in Table.~\ref{tab:quantitative_scand}, our method closely aligns with human preference, demonstrating socially-aware behaviours across diverse scenarios.
In comparison, path-etiquette uses the same VLM model as ours to perform semantic reasoning, but the resulting trajectories are not as natural due to poor sampling quality.
While CityWalker shows competitive performance on both metrics, it tends to generate highly homogeneous motions, a bias likely stems from its large-scale training.
We also perform an ablation study where we removed each of the key modules in our framework.
The results show that traversability identification and social scoring are the crucial components that contribute to social context understanding.
Notice that our method still performs strongly without the safeguard.
This is because the human operator was fully aware of the surroundings during data collection, therefore did not perform any urgent avoidance.
However, we will show in the qualitative analysis that the safeguard is an important feature during autonomous navigation.

\begin{figure}[h]
    \centering
    \includegraphics[width=\linewidth]{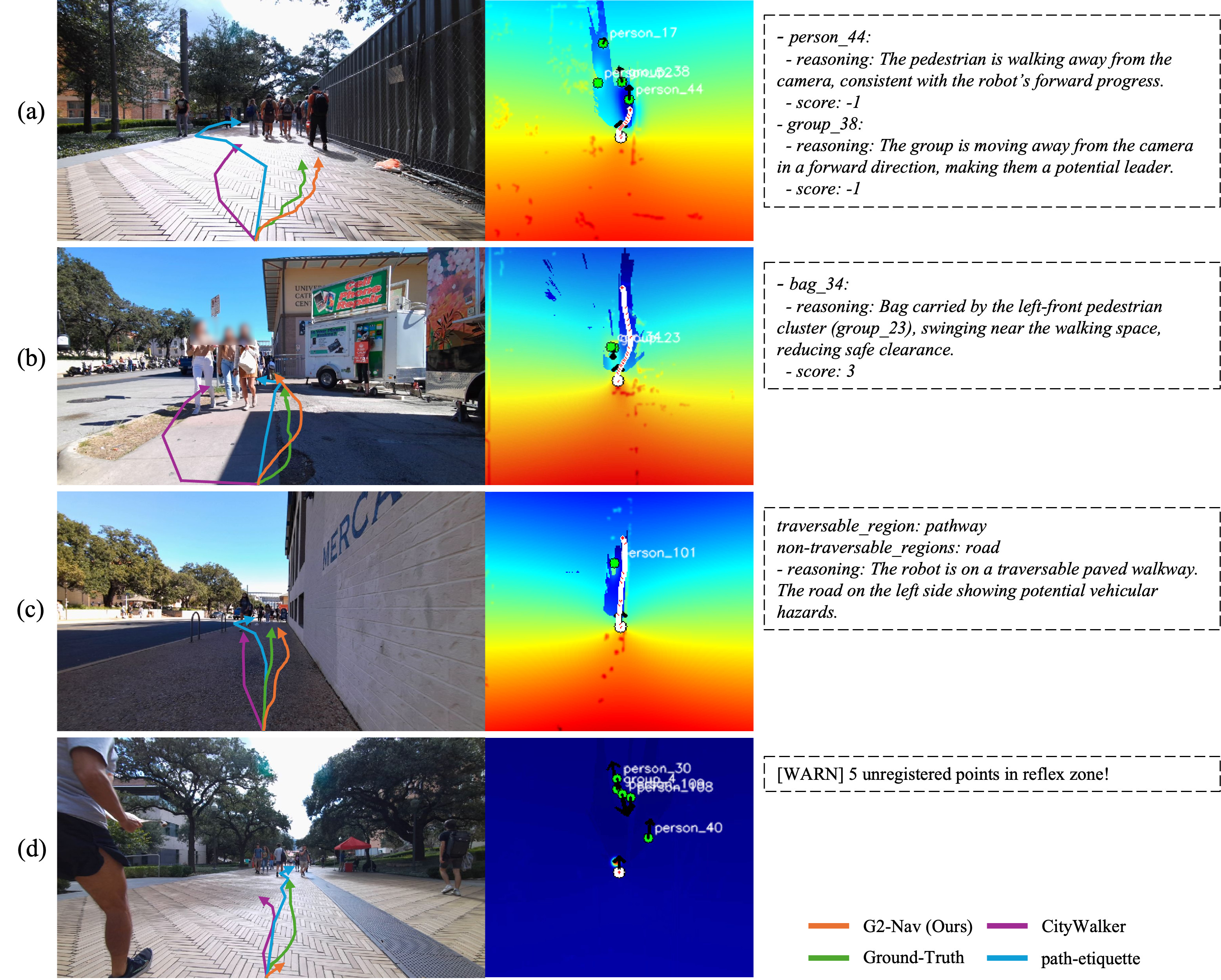}
    \caption{Qualitative results under recorded setting. On the left, we project ground-truth trajectories from the next 5 seconds onto the current frame, overlayed by projected trajectories planned by our method and baselines. In the middle, we show the generated BEV costmap and derived trajectory. On the right, we show the key VLM response snippet in (a) - (c), and the reflex zone warning in (d).}
    \label{fig:scand_qualitative}
\end{figure}

\noindent\textbf{Qualitative Analysis.}
We highlight the socially responsible and safety-assured behaviours of our method through qualitative comparisons.
As shown in Fig.~\ref{fig:scand_qualitative}(a), instead of trying to cut in front of the group and potentially result in collisions with approaching pedestrians, VLM suggests that the robot should follow the traffic, which aligns with the decision from the human operator.
In Fig.~\ref{fig:scand_qualitative}(b), the robot benefits from the unique information about the swinging bag, ensuring a larger safety clearance.
In a pathway example shown in Fig.~\ref{fig:scand_qualitative}(c), both baselines steer to the left to avoid the wall, but it is in fact a safer decision to stay on the narrow pathway.
We demonstrate the safeguard feature in Fig.~\ref{fig:scand_qualitative}(d) when a person runs past the robot in close distance.
Although the human operator might already know that the person is coming and did not respond, our method suggests a safety brake, which is crucial when interacting with unpredictable humans in the real-world.

\begin{table}[h]
    \centering
    

    \begin{tabular}{l cccc}
        \toprule
        \textbf{Method} & \textbf{SVT ($s$) $\downarrow$} & \textbf{T\textsubscript{avg} ($s$) $\downarrow$} & \textbf{D\textsubscript{avg} ($m$) $\downarrow$} & \textbf{Intervention (Safety / Bound.) $\downarrow$}\\ 
        \midrule
        DWA~\cite{fox1997dynamic}      & 11.9 & 392 & 181.6 & \underline{2} / \textbf{0} \\
        SF~\cite{helbing1995social}      & \textbf{4.4} & 352 & \textbf{165.4} & 3 / 13 \\
        PAP~\cite{liao2025following}      & 8.1 & \underline{322} & 180.9 & 6 / \underline{7} \\
        CityWalker~\cite{liu2025citywalker}      & 16.6 & 403 & 179.1 & 15 / 9 \\
        path-etiquette~\cite{fang2026obstacles}  & 10.2 & 518 & 198.4 & 10 / 11 \\
        \midrule
        \textbf{G2-Nav} (ours)   & \underline{6.9} & \textbf{278} & \underline{172.3} & \textbf{0} / \textbf{0} \\
        \bottomrule
    \end{tabular}

    \vspace{+5pt}
    \caption{Quantitative comparisons under interactive setting. Results are averaged from 6 trials, except the intervention counts, which report the total number of safety and boundary pauses by the human supervisor over all 6 trials. The best performance is in \textbf{bold} and the second-best is \underline{underlined}. Travel time may increase for the same distance if the robot deadlocks or needs to accelerate from the supervisor intervention pause.}
    \label{tab:quantitative_real}
\end{table}

\subsection{Interactive Setting}
\noindent\textbf{Quantitative Analysis.}
As shown in Table.~\ref{tab:quantitative_real}, our method demonstrates strong safety awareness without compromising navigation efficiency.
SF is a competitive baseline, but it maintains safety clearance by either staying on the middle of the road, which is not a socially responsible behaviour, 
or attempting to steer into the car park, requiring frequent supervisor rescue.
PAP uses SF as its base planner and can move more smoothly with the crowd flow.
However, it often misidentifies human leaders due to tracking noises, which lead to confusing oscillatory motion.
DWA does not plan forward and therefore often makes abrupt stops to avoid obstacles, or gets trapped in a corner.
Citywalker can generate fundamentally sound trajectories, but it is less sensitive to spatial details such as small fixtures on the wall, which then lead to dangerously close contact.
Path-etiquette relies on the VLM selected reference trajectory which is updated at a lower frequency.
This results in the robot missing critical real-time information.
In contrast, our costmap is updated asynchronously, integrating social context upon every VLM inference cycle.
This makes G2-Nav a robust framework suitable for real-world deployment.

\begin{figure}[t]
    \centering
    \includegraphics[width=\linewidth]{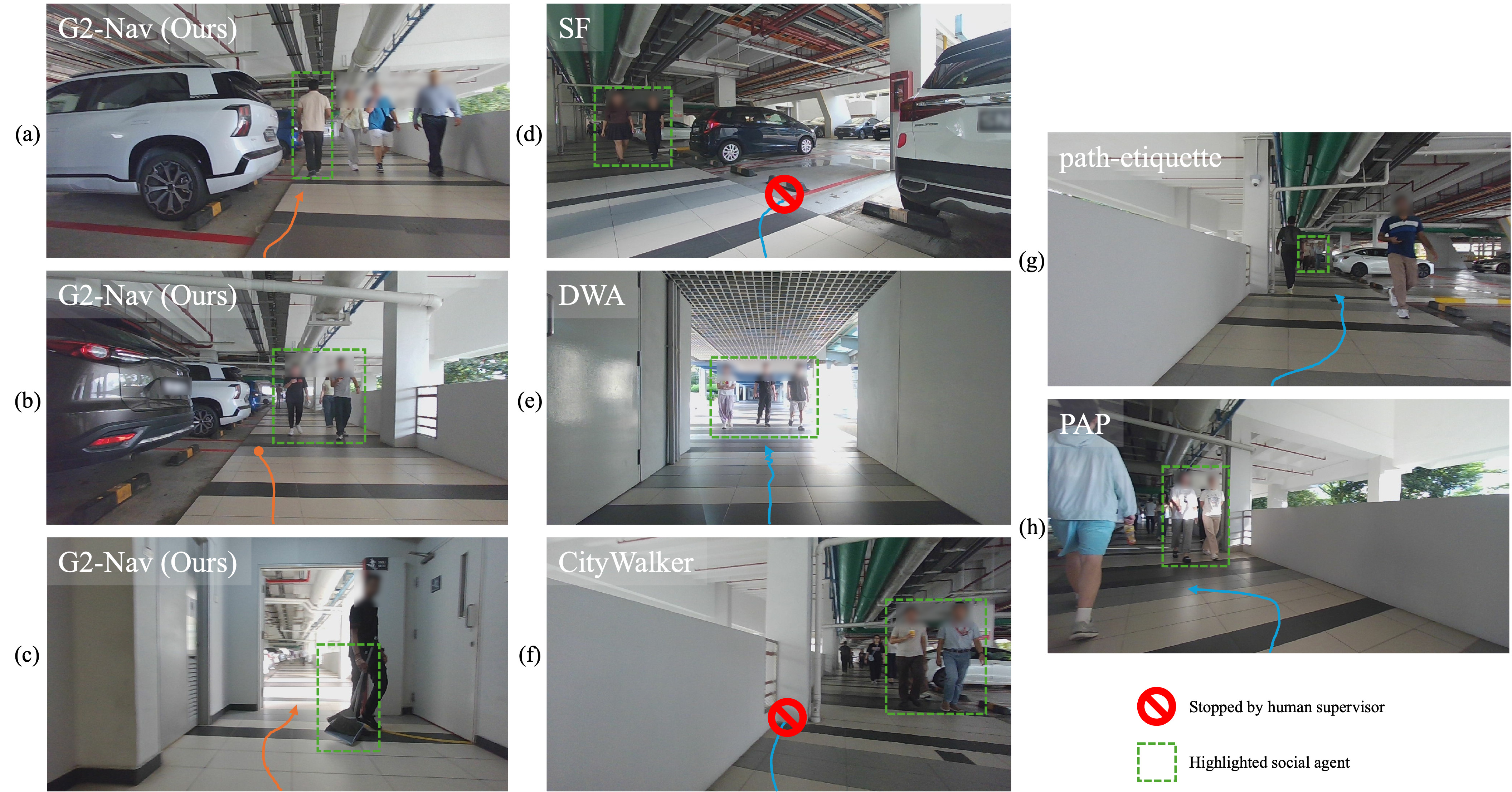}
    \caption{Qualitative results under interactive setting. We project the robot trajectories from the next 5 seconds onto the current frame. We showcase 3 socially responsible behaviours of our method and a failure example for each baseline. Experiment recordings can be found in supplementary video.}
    \label{fig:real_qualitative}
\end{figure}

\noindent\textbf{Qualitative Analysis.}
We observe socially responsible and safety-aware behaviours similar to what we have highlighted in the recorded setting, including following human leader in Fig.~\ref{fig:real_qualitative}(a), giving way to approaching group in Fig.~\ref{fig:real_qualitative}(b), and avoiding the swinging tool in Fig.~\ref{fig:real_qualitative}(c).
To supplement what have been discussed in the quantitative analysis, we show in Fig.~\ref{fig:real_qualitative}(d) where SF attempts to steer into the car park to avoid approaching pedestrians.
In Fig.~\ref{fig:real_qualitative}(e), DWA abruptly stops in front of the group without giving way early, which can lead to human discomfort.
In Fig.~\ref{fig:real_qualitative}(f), CityWalker avoids the group but overlooks the protruding pipe.
Path-etiquette plans a valid path in Fig.~\ref{fig:real_qualitative}(g), but without real-time update, this path will lead to collisions in the next few frames.
In Fig.~\ref{fig:real_qualitative}(h), PAP makes a confusing orthogonal turn, likely because the approaching group was mistaken as the leader and later corrected with updated tracking results.
Our upstream verification step is particularly designed to avoid such confusion using VLM spatial reasoning capability.

\section{Conclusion}

We present G2-Nav, a social navigation framework that grounds VLM semantic reasoning into reliable costmap-based robot behaviours.
At its core is a vision-language costmap that translates social interaction analysis into bounded navigation costs, paired with an upstream verification mechanism and a safety layer to ensure robustness. 
Extensive experiments with real-world social interactions demonstrate that G2-Nav promotes safe, efficient, and socially responsible robot behaviours.

\section{Limitations and Future Work}

G2-Nav inherits the failure modes of its individual components.
SAM could be confused by ground patterns, producing inaccurate mask which leads to trajectories outside traversable regions.
The VLM may hallucinate by assigning the same score to all objects, leading to over-conservative behaviours.
Future works can remodel the multi-step open-set perception pipeline into a carefully trained network, and finetune or distill the VLM for more reliable performance.
The current safeguard layer only checks for unregistered nearby point clouds.
More safeguard layers can be applied to the vision-language costmap, such as using the Control Barrier Function~\cite{ravichandran2026contextual} for safety guarantee, or techniques in dynamic systems to avoid trajectories being trapped in local minima~\cite{sindhwani2018learning}.



\clearpage
\acknowledgments{
  This research was supported by Ministry of Education, Singapore, under AcRF TIER 1 Grant RG64/23, 
  and the Centre for Advanced Robotics Technology Innovation (CARTIN) funded by the National Research Foundation of Singapore under its Medium-Sized Centre Scheme.
  The authors would like to thank the reviewers for their constructive feedback that helped improve this manuscript.
}


\bibliography{references}  

\clearpage 

\appendix
\section*{Appendix}
\renewcommand{\thesubsection}{\Alph{subsection}} 

\subsection{Upstream Verification Example}
\begin{figure} [h]
    \centering
    \includegraphics[width=\linewidth]{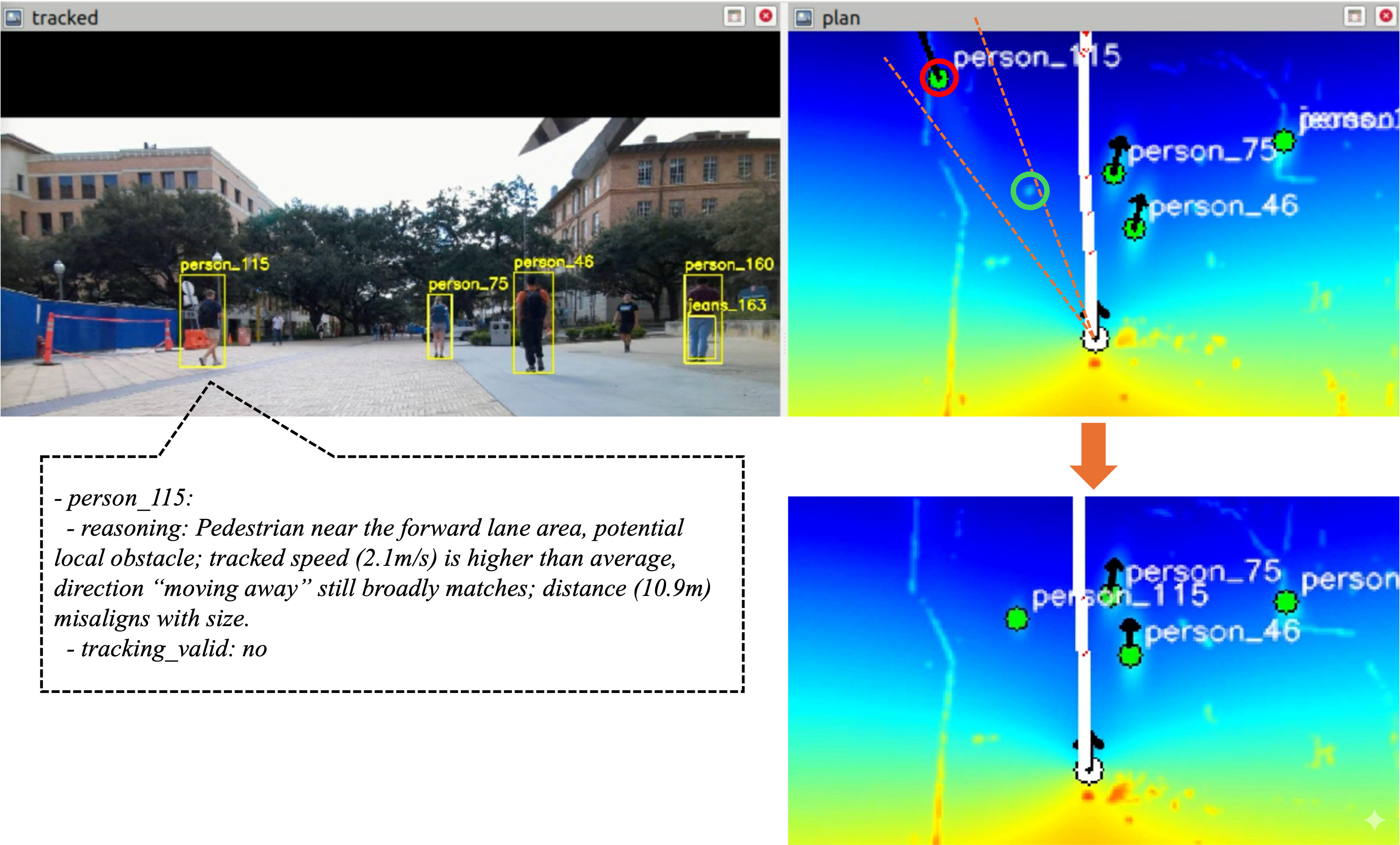}
    \caption{Example of the upstream verification step. The VLM response snippet suggests that both the motion and depth information about person\_115 mismatch with visual observations. We therefore perform depth re-registration in the same projected window (orange dotted lines) while filtering out the false background cluster (red circle), and successfully identify the correct cluster for person\_115 (green circle).}
    \label{fig:verification}
\end{figure}
The VLM verifies whether the tracked object's depth and heading align with visual observations.
As illustrated in Fig.~\ref{fig:verification}, the initial depth registration process mis-assigned background features to person\_115, which subsequently results in inaccurate velocity estimation.
This discrepancy is successfully detected by the VLM.
We therefore assign person\_115 a higher social score to account for its uncertain motion, and perform depth re-registration by searching in the same projected window while filtering out the previous false cluster.
This correction will be applied to person\_115 until the VLM verifies the latest tracking results in the subsequent inference cycle.

\subsection{Reflex Zone Definition}
We define a dynamic reflex zone along the robot's predicted trajectory, and treat unregistered LiDAR points within this reflex zone as immediate threats.
Based on the robot's current linear velocity $v$ and angular velocity $\omega$, the reflex zone $\mathcal{Z}$ is either a rectangular box or a curved arc, defined in the robot's local frame as follows:
\begin{equation}
    \mathcal{Z} = 
    \begin{cases} 
    \left\{ (x, y) \;\Bigg|\; 0 < x < vT, \; |y| < \frac{W}{2} \right\}, & \text{if } |\omega| < \epsilon, \\
    \left\{ (x, y) \;\Bigg|\; 0 < \theta < |\omega|T, \; \Big| \sqrt{x^2 + (y - \frac{v}{\omega})^2} - |\frac{v}{\omega}| \Big| < \frac{W}{2} \right\}, & \text{if } |\omega| \ge \epsilon, 
    \end{cases}
\end{equation}
where $\theta = \text{arctan2}(x, |\frac{v}{\omega}| - y \cdot \text{sgn}(\omega))$ is the swept angle, $T$ is the time-to-collision window, $W$ is the zone width, and $\epsilon$ is the minimum angular velocity threshold to avoid numerical instability.
In the experiments, we set $T$, $W$, and $\epsilon$ to 1.5s, 0.8m, and 0.05 respectively.
The reflex zone was activated 18 times in SCAND recorded setting ($\sim$3/h), compared to 10 times in real-world interactive setting ($\sim$20/h).
This shows that the safety layer is much more active on an autonomous robot with only onboard sensors.

\subsection{Latency Ablation Study}
\begin{wrapfigure}{r}{0.5\textwidth}
    \vspace{-10pt}
    \centering
    \includegraphics[width=\linewidth]{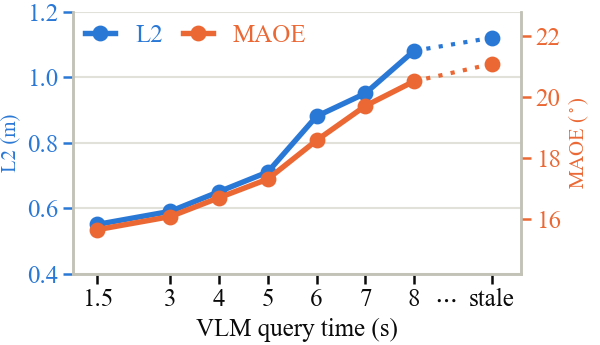}
    \vspace{-15pt}
    \caption{Latency ablation in recorded setting.}
    \label{fig:latency}
    \vspace{-10pt}
\end{wrapfigure}
As shown in Fig.~\ref{fig:latency}, we perform a latency ablation study under the recorded setting, where VLM response is held to mimic different query times.
The stale output is the same as ``Ours w/o scoring'' in Table.1 of the paper, where all objects are given default scores of 3.
Both metrics degrade gracefully toward the stale bound, with 82\% (L2) and 81\% (MAOE) of VLM's benefit retained at the deployed 4s latency in the interactive experiment.
Any mis-registered agent's true LiDAR points still occupy raw occupancy layer, so collision avoidance is not compromised.
VLM verification then restores semantics at the next cycle.
Increasing latency reduces freshness of semantic guidance, quantified by this ablation study.

\subsection{VLM Prompt}
The full prompt is shown in Prompt~\ref{lst:prompt}.
Three consecutive camera frames spaced 1s apart are attached as images before the text.
The placeholders \texttt{\{objects\_text\}}, \texttt{\{region\_list\}}, and \texttt{\{traversable\_region\}} are filled at runtime.
The VLM response is parsed with regular expressions to extract the traversable region, the new non-traversable regions, and the social score and verification flags of each object.
More details on calling the VLM and post-processing the response can be found in the code repository.

\begin{lstlisting}[caption={Full prompt for the VLM component in the proposed G2-Nav framework.}, label={lst:prompt}]
You are an autonomous robot navigating a campus environment, currently moving forward. Your goal is safe and efficient forward navigation.
The attached images are from your onboard camera, spaced 1 second apart (Frame 1 to Frame 3). Frame 3 is your current view.

Detected objects in Frame 3 are listed below, in the format `- label_ID detection_confidence: bbox=(xmin, ymin, xmax, ymax), tracking_distance: [X.Xm], tracking_motion: [Speed and Direction]`:
{objects_text}

--- REGION IDENTIFICATION TASK ---
Currently known non-traversable regions: {region_list}
Current traversable region: '{traversable_region}'

1. Check if any objects in the list above describe a region (e.g., grass, road, sidewalk, floor, ground) rather than a physical obstacle.
2. Based on the images, check if the 'Current traversable region' is still valid and safe for forward navigation. If it is empty or no longer valid, pick a new traversable region from the newly identified regions.
3. Any newly identified regions from the object list that are NOT the traversable region should be marked as new non-traversable regions.

--- OBSTACLE EVALUATION TASK ---
Identify objects from the list that may influence your navigation. Consider ALL types of objects: dynamic (e.g., pedestrians, bicycles, cars) and static (e.g., traffic signs, safety marks, obstacles).
For each selected object, decide whether the robot should AVOID it as an obstacle, FOLLOW it as a leader, or IGNORE it.
Additionally, verify if the provided tracking_motion matches your visual observation of the object's absolute motion relative to the ground.

Rules:
1. STRICTLY use the exact `label_ID` (e.g., `bicycle_160`) from the provided list. Do NOT invent new objects or alter the IDs.
2. For dynamic objects, infer their relative motion (e.g., approaching, crossing, receding) by comparing their positions across Frame 1, 2, and 3.
3. Leader Candidates: You may assign a score of -1 to follow an object IF it is moving in a similar forward direction and it is safe/beneficial to follow. Do NOT force finding a leader if none are suitable (e.g., crowds walking against you).
4. Ignore Irrelevant Objects: If an object is irrelevant, or it belongs to a tracked human (e.g., a backpack, shoe, clothing) and it's not affecting the robot directly, it should be marked to IGNORE (score 0). Objects identified as regions in the Region Identification Task should NOT be evaluated here as obstacles.
5. Tracking Check: Evaluate if the tracking_motion makes sense. If an object is visually static but has a high speed, or its tracked direction clearly contradicts visual evidence, flag it as tracking_valid: no. Note: average pedestrian walking speed is 1.4m/s; if a human's speed is significantly higher than this, you need to carefully examine if it's valid.
6. Depth Check: Evaluate if the tracking_distance matches the visual scale of the object. If the object appears close but has a large tracking_distance, flag it as depth_valid: no.
7. Keep your analysis brief to minimize computational load, but think step-by-step before assigning a score.

Answer strictly in the following format:
Before evaluating obstacles, you MUST output the following two lines exactly:
TRAVERSABLE_REGION: <label>
NEW_NON_TRAVERSABLE_REGIONS: [<label1>, <label2>, ...]

Then, for each object that is NOT a region, provide:
- label_ID:
  - status: [Briefly state if static or moving. If moving, describe relative motion across frames]
  - reasoning: [1 short sentence explaining why it is important, whether visual motion matches tracking_motion, AND whether visual scale matches tracking_distance]
  - tracking_valid: [yes/no] (Answer 'yes' if they match. Answer 'no' if they strongly contradict)
  - depth_valid: [yes/no] (Answer 'no' if visual scale strongly contradicts tracking_distance)
  - score: [-1~5] (-1=follow as leader, 0=ignore, 1~5=avoid as obstacle where 1=least important, 5=safety-critical)
(leave a blank line between each object)
\end{lstlisting}

\end{document}